\documentclass[conference]{IEEEtran}
\IEEEoverridecommandlockouts
\usepackage{cite}
\usepackage{amsmath,amssymb,amsfonts}
\usepackage{algorithmic}
\usepackage{graphicx}
\usepackage{textcomp}
\usepackage{xcolor}
\usepackage{hyperref}

\usepackage{booktabs}
\usepackage{adjustbox}
\usepackage{multirow}
\usepackage{hhline}
\usepackage{makecell}

\def\BibTeX{{\rm B\kern-.05em{\sc i\kern-.025em b}\kern-.08em
    T\kern-.1667em\lower.7ex\hbox{E}\kern-.125emX}}
\begin{document}

\title{Inclusive normalization of face images to passport format\\
}

\author{\IEEEauthorblockN{1\textsuperscript{st} Hongliu CAO}
\IEEEauthorblockA{\textit{Amadeus SAS}\\ \textit{hongliu.cao@amadeus.com}}
\and
\IEEEauthorblockN{2\textsuperscript{nd} Minh Nhat Do}
\IEEEauthorblockA{\textit{Amadeus SAS} }
\and
\IEEEauthorblockN{3\textsuperscript{rd} Alexis Ravanel}
\IEEEauthorblockA{\textit{Amadeus SAS} }
\and
\IEEEauthorblockN{4\textsuperscript{th} Eoin Thomas }
\IEEEauthorblockA{\textit{Amadeus SAS} }
}

\maketitle

\begin{abstract}
Face recognition has been used more and more in real world applications in recent years. However, when the skin color bias is coupled with intra-personal variations like harsh illumination, the face recognition task is more likely to fail, even during human inspection. Face normalization methods try to deal with such challenges by removing intra-personal variations from an input image while keeping the identity the same. However, most face normalization methods can only remove one or two variations and ignore dataset biases such as skin color bias. The outputs of many face normalization methods are also not realistic to human observers. In this work, a style based face normalization model (StyleFNM) is proposed to remove most intra-personal variations including large changes in pose, bad or harsh illumination, low resolution, blur, facial expressions, and accessories like sunglasses among others. The dataset bias is also dealt with in this paper by controlling a pretrained GAN to generate a balanced dataset of passport-like images. The experimental results show that StyleFNM can generate more realistic outputs and can improve significantly the accuracy and fairness of face recognition systems.  
\end{abstract}


\section{Introduction}

With the development of deep learning algorithms, computer vision applications have received great success in recent years. Among which, facial recognition has gained more and more attention due to its wide application in both 
unconstrained environment  (e.g.  certain surveillance systems, photo album management in social networks, human machine interaction, and digital entertainment) \cite{klare2015pushing, zhao2020recognizing, han2019asymmetric} and in constrained environment (e.g. automatic check-in or baggage drop at the airport) \cite{adjabi2020past, masi2018deep}. However, there are still a lot of challenges in face recognition systems, especially in dealing with image diversity, generalization ability and fairness. In real world applications, the enrolled faces in the gallery are usually frontal standard photos such as passport-like or other ID-like pictures, while the probe face pictures can be very diverse in terms of poses, bad or harsh illumination, low resolution, blur, facial expressions, accessories like sunglasses and so on \cite{tu2021joint}. When bad illumination couples with skin color bias, the face recognition problem can be very difficult even for human inspections.

Generative Adversarial Networks (GANs) have achieved great performance in terms of high resolution face image generation in recent years especially with the StyleGAN series \cite{karras2019style, karras2020analyzing, karras2020training}. Image to image translation tasks have also been widely explored using GANs such as CycleGAN \cite{zhu2017unpaired} or StarGAN \cite{choi2018stargan, choi2020stargan}. With these two types of GANs, the previously mentioned challenges in face recognition can be mitigated by: 1. using GANs to generate training data to deal with dataset bias and to improve the fairness, 2. using image to image translation GANs to normalize the input face image to a standard, easy to recognize format for both machine learning systems and human beings. 

In the literature, there are many solutions trying to use image-to-image translation GANs to normalize face images, but most state-of-the-art solutions have three common limitations. 
The first one is that existing face image normalization methods typically tackle the previous mentioned challenges (large pose, harsh illumination, facial expression, accessories and so on) individually. In \cite{tu2021joint}, the authors found out that many existing solutions fail on multi-factor cases. 
The second common limitation is the dataset bias. Training an image-to-image translation GAN needs both non-normal images (different variations such as large pose or bad illumination as input) and normal images (the standard format we want as output), some solutions even require the non-normal and normal image datasets to be paired. Most studies have difficulties finding a good training dataset to fulfill these requirements and ignore the other dataset bias such as gender and skin color bias. 
The third common limitation is that even though existing face normalization methods can improve the face recognition performance to a certain extent, the output face images are not realistic and difficult to recognize for human observers (some examples shown in Figure \ref{fig:2}). 
The last common limitation is the inference speed. In real-world applications, the face image normalization module is ran as a pre-processing step for facial recognition system, which needs to be fast. However, many solutions have such complex models that the inference speed is slow.   

In this work, we propose a framework to deal with the above mentioned limitations of the state-of-the-art solutions to make the face recognition task easier for both human inspectors and machines. The main contributions of this work are:
\begin{itemize}
    \item A face normalization framework is proposed to normalize most variations using a single model. The proposed model is the first in the literature that can normalize a face image input to a passport-like standard image. Our model can remove sunglasses, normalize the original harsh illumination along with the background to standard passport-image-like frontal light with white background, normalize the original pose and facial expression to frontal standard passport-image-like pose and neutral expression, remove effects like blur and so on, while keeping the original identity.   
    \item The proposed face normalization framework deals with dataset bias by controlling pretrained GANs to generate a normal dataset with  balanced gender and skin color distribution, which improves significantly the generated output for all targeted ethnicities. 
    \item The proposed framework can generate more realistic images and improve significantly the performance of baseline face recognition system compared to other face normalization methods in the literature. The proposed framework also has very good generalization performance on other datasets in the context of the presence of domain gaps.   
    \item The proposed framework takes generation speed at inference time into consideration and achieves  fast inference speed to ensure its applicability. 
\end{itemize}

\section{Related works}
Since first introduced by Goodfellow et al. \cite{goodfellow2020generative}, GANs have achieved great success in many fields including realistic image generation, text generation, super resolution, image inpainting, image manipulation, etc. The general goal of GAN is to allow the generator to learn a distribution that matches the
real data distribution via an adversarial loss from the discriminator, which distinguishes the generated data from real data \cite{han2019asymmetric}. In this section, the literature of both GANs for image generation and GANs for face normalization are introduced. 

\subsection{GANs for image generation}
The resolution and quality of synthetic photo-realistic face images generated by GANs have improved in the past years \cite{arjovsky2017wasserstein, brock2018large, miyato2018spectral, karras2017progressive}. 
The generators of traditional GANs (e.g. Progressive GAN \cite{karras2017progressive}) take a random noise vector as input and transform it into a realistic image with convolutional layers.   
Among the most recent works, style based GANs are the most successful ones in terms of generating realistic face images with less entangled latent spaces. Instead of using the random noise vector as input to the generator, StyleGAN series have a mapping network to map the random noise vector in Z space to the latent W space. Then the latent vector W is used to generate style vectors that are injected into different layers of the generator. StyleGAN \cite{karras2019style}  uses adaptive instance normalization to inject the styles while StyleGAN2 \cite{karras2020analyzing} realizes it by modulating the weights of
the convolution kernels. The StyleGAN series are one of the best performing GANs in the literature in terms of generating high quality high resolution photo-realistic face images. Compared to other state-of-the-art GANs,  StyleGAN series also have the advantage of having a less entangled latent space which makes it easier to manipulate for the  generation of the desired face variations.   

\subsection{GANs for face normalization}
The main objective of face normalization in the context of face recognition is to remove the possible variations or combinations of multiple variations (including large poses, bad illumination, heterogeneous background, sunglasses, blur, etc.) to get a standard passport-like images to facilitate the face recognition task.  
In recent years, many research studies have tried to use GANs for face normalization, however the majority of them only normalize one or two variations. For example, many works focus on pose normalization to normalize the input faces with different poses to a standard frontal pose \cite{tran2017disentangled, zhao2018towards, hu2018pose}. Some solutions even add illumination variations or glasses to the output image while normalizing the pose. Beyond pose variation, appearance variation caused by illumination is another major problem that remains unresolved in face recognition \cite{han2019asymmetric}. Many studies in the literature also try to use GANs for illumination normalization such as in \cite{ma2018face, zhang2019gan}, however they are limited to the illumination types in the training data.    

In the literature, there are several studies attempts to deal with multiple variations at the same time. For example, the researchers in \cite{nagano2019deep} proposed the deep face normalization framework to normalize the input face images. However, the solution involves cascading multiple models together to realize the task rather than building a single model, leading to some limitations. Firstly, the latter model inputs depend on the previous model outputs which would accumulate the errors of all previous models and make the outputs unstable. Secondly, if a new variation like sunglasses needs to be removed, a new model needs to be added to the model chain, which is difficult to generalize. Thirdly, the face normalization process is a pre-processing step which needs to be fast, yet model combination approaches may lead to additional latency and processing power requirements.

\begin{figure}
    \centering
    \includegraphics[width=0.49\textwidth]{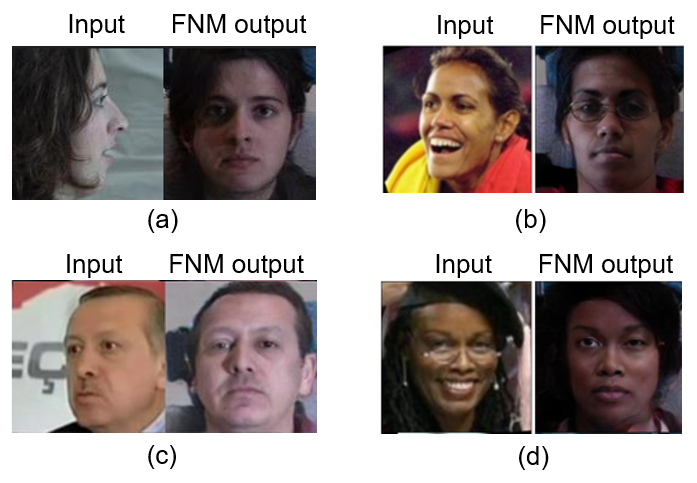}
    \caption{Examples of FNM outputs from different inputs.}
    \label{fig:2}
\end{figure}

In \cite{qian2019unsupervised}, the authors proposed Face Normalization Model (FNM) to normalize both pose and facial expression at the same time. FNM takes advantage of the face expert network as prior knowledge to encode the input images and builds a generator afterwards with three losses. 
In this work, FNM is chosen as the baseline solution due to the following reasons. Firstly, the main advantage of FNM is that no paired datasets are needed so that multiple datasets can be combined together to train the model.  Secondly, FNM only uses one generator which has the potential to speed up real world applications. However, there are several limitations of FNM, some of which are illustrated in Figure \ref{fig:2}. Firstly, FNM has issues with some variations such as illumination. From Figure \ref{fig:2} (a), it can be seen that FNM adds random illuminations to the outputs. This phenomenon can also be observed from Figure \ref{fig:2} (b) - (d). Secondly, FNM outputs are inconsistent with the variation of glasses. In Figure \ref{fig:2} (b), FNM adds glasses to the normalized output. From the same figure, we can see that FNM performs poorly with different skin colors. The example of Figure \ref{fig:2} (c) shows that the pose variation is not well normalized. From the example in Figure \ref{fig:2} (d), we can see a problem with the generation quality of FNM (asymmetry and unrealistic eyes). To deal with these issues, we propose improvements to the generator and discriminator, as well as a reduction in dataset bias, in the next section.

\section{Approach}
\begin{figure}
    \centering
    \includegraphics[width=0.45\textwidth]{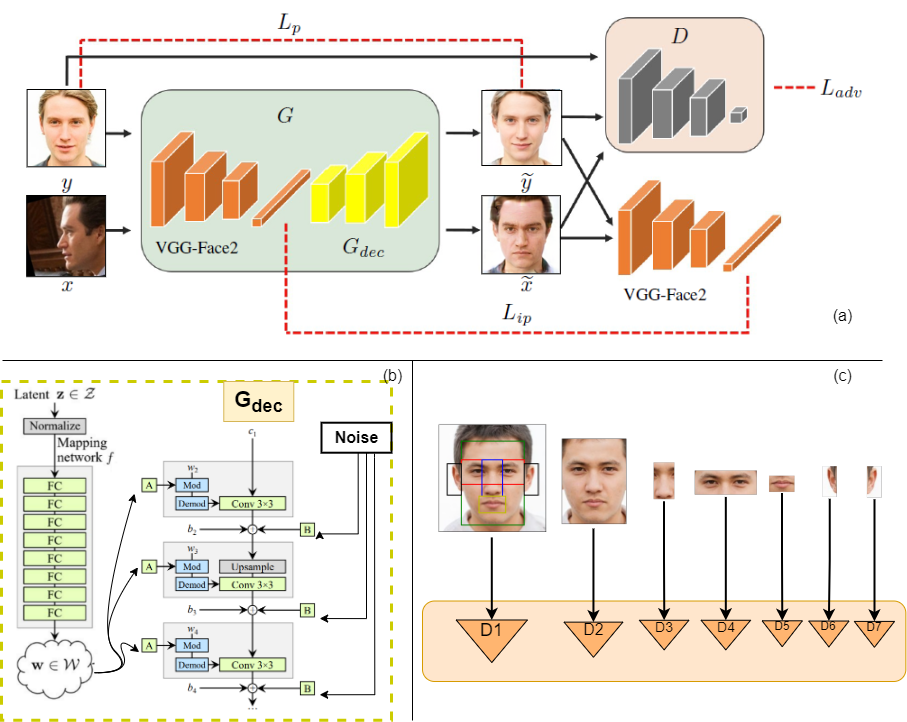}
    \caption{The general framework of StyleFNM: (a) overall structure of StyleFNM, (b) the generator structure, (c) the discriminators structure.  }
    \label{fig:framework}
\end{figure}

\subsection{The overall framework}
The overall proposed framework StyleFNM is shown in Figure \ref{fig:framework}. Similar to the work of FNM, the unpaired data training and face expert network are used. The main changes lie in the normal data set generation to reduce the dataset bias, the generator, the discriminator design and the loss function. 

As shown in Figure \ref{fig:framework} (a), the input images are firstly fed into an encoder, then the encoded information is fed into the generator/decoder. As the only information we want to keep from the input is the identity information, a pretrained VGG-Face2 \cite{cao2018vggface2} network is used as the encoder to take advantage of the prior knowledge of the face expert. From the examples in Figure \ref{fig:2}, we conclude in the previous section that the generator of FNM is not powerful enough to keep the identity and generate photo-realistic face images. In this work, we propose to use the generator of StyleGAN2 \cite{karras2020analyzing} instead (as shown in Figure \ref{fig:framework} (b)). To guarantee the fast speed of both training and testing, the output size of the generator is fixed as 128x128 instead of 224x224 (FNM output size). The generated images are passed through the discriminator along with the real normal images.

\subsection{The generator}
As shown in Figure \ref{fig:framework} (a), the generator contains two parts: the pretrained VGGFace2 is used as encoder $G_{enc}$ to extract identity information, while StyleGAN2 is used as decoder $G_{dec}$  to generate passport like images while keeping the identity information. Two datasets are needed to train the generator: non-normal dataset (face images with all kinds of variations) as input and normal dataset (passport-like standard images) as output. The main task is to transform the non-normal images to normal images. To deal with the real world dataset lacking limitations, the non-normal and normal datasets are not paired in this work.

Let $x \in \mathbb{R}^{H\times W\times C} $ denote one face image from the non-normal dataset and $y \in \mathbb{R}^{H\times W\times C} $ denote one face image from the normal dataset. When passing $x$ from the non-normal dataset into the generator, we expect the generator output $\Tilde{x}$ to be the normalized version of $x$. But when passing $y$ from the normal dataset, the expected output from the generator $\Tilde{y}$ should be the same as $y$. Hence, for the inputs from normal image dataset, we can use the pixel wise loss $L_p$ to enforce the generator to learn from such supervised information:

\begin{equation}\label{lp}
    L_p = \frac{1}{W \times H \times C} \sum_{w,h,c}^{W,H,C} |y_{w,h,c} - \tilde{y}_{w,h,c}|
\end{equation}

For input images from non-normal dataset, pixel-wise loss is not suitable anymore as we have no paired data. Instead, the identity perception loss $L_{ip}$ is used to ensure that the generated image and the input image have the same identity:

\begin{equation}\label{lip}
    L_{ip} = ||G_{enc}(x) - G_{enc}(\tilde{x})||^2_2  + ||G_{enc}(y) - G_{enc}(\tilde{y})||^2_2
\end{equation}

In Figure \ref{fig:2} (c), we can see that the pose variation is not always well normalized by FNM. Figure \ref{fig:2} (d) shows that some of the FNM generated images are less realistic due to differences between the left and right sides of the face. To deal with these issues, we introduce an extra symmetric loss term $L_{sym}$ for both normal inputs and non-normal inputs:
\begin{equation}\label{lsym}
    L_{sym} = \frac{\sum_{w,h,c}^{W,H,C} |\tilde{x}_{w,h,c} - \tilde{x}^{hf}_{{w,h,c}}| + \sum_{w,h,c}^{W,H,C} |\tilde{y}_{w,h,c} - \tilde{y}^{hf}_{{w,h,c}}|}{2\times w\times h \times c}
\end{equation}

where $\Tilde{x}^{hf}$ means the horizontal flip of $\Tilde{x}$. With the symmetric loss $L_{sym}$, we can penalize the generator when the outputs have non-frontal illumination, not well normalized pose as well as other unrealistic asymmetry of left and right face. 

\subsection{The discriminator}
In this work, 7 discriminators are constructed. Each of these discriminators pays attention to a different region of the face to produce photo realistic face images of high quality.
As the normal image dataset is aligned, we expect the generated images to be aligned too. Hence, all the attention regions are pre-fixed instead of being learned. As shown in Figure \ref{fig:framework} (c), the discriminators' attention region are: full image, face region only, nose region, eye region, mouth region, left ear region and right ear region respectively. 
The eye region in this work is larger than that of FNM. The main reason is that we want to remove accessories like sunglasses from the input image. If we only focus on a small eye region, the rest of the sunglasses (the frames) may stay on the face. 
We also specifically add two ear regions to deal with the unrealistic ears frequently generated by FNM. The adversarial loss $L_{adv}$ of the discriminators is:

\begin{equation}\label{ladv}
    L_{adv} = \sum_{k=1}^{7} D_k(\tilde{x}_k) + \sum_{k=1}^{7} D_k(\tilde{y}_k) - \sum_{k=1}^{7} D_k(y_k) 
\end{equation}

The final loss of the generator $L_G$ is:
\begin{equation}\label{lgen}
    L_{Gen} =  L_{adv} + \lambda_1L_{ip} + \lambda_2L_p + \lambda_3L_{sym}
\end{equation}

Similar to the StyleGAN2, we also apply the Lazy Regularization and Path length regularization \cite{karras2020analyzing} to improve the quality of generated image and to stabilize the training process of the model.

\subsection{Normal dataset generation with pretrained GANs}
As analyzed in Section 2, two of the main limitations of the current state-of-the-art methods in face normalization are the generator capacity and dataset bias. In the previous subsection, improvements to the generator, discriminator and loss functions are proposed. Here, we propose a methodology to reduce dataset bias. 

The first challenge in terms of dataset bias is that there are no large datasets of real normal pictures (passport-like standard images) available. In the literature, many studies use the CMU Multi-PIE dataset \cite{gross2010multi}. The main problem of using Multi-PIE dataset as normal dataset is that the  sample size of frontal illumination with neutral pose is very small (150 images in setting 1).  

Due to legal and privacy reason, it is unfeasible to have real passport images as the normal dataset. In this work, we propose to control the pretrained GANs to generate the passport-like normal dataset. The pretrained GAN we selected is the StyleGAN2-Ada due to the high quality of its generated images. StyleGAN2-Ada trained on FFHQ dataset also has no large pose variations nor harsh illuminations. The latent W space of the StyleGAN series also has the advantage of being more disentangled, which makes it easier to control the GAN to generate the variations of images we require.  In the work of InterFaceGAN \cite{shen2020interfacegan}, the authors found out that GANs learn various semantics in some linear subspaces of the latent space. Once these subspaces are identified, we can manipulate the pretrained GAN to generate the desired variations such as gender, age, expression, etc. 
\begin{figure}
    \centering
    \includegraphics[width=0.4\textwidth]{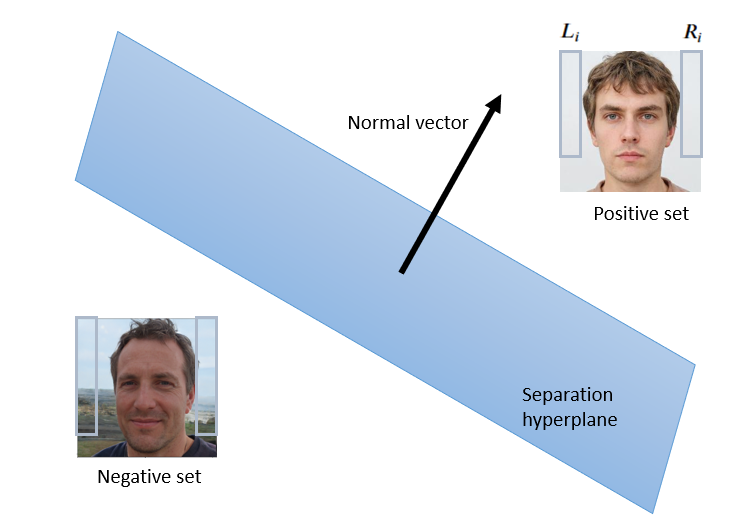}
    \caption{The procedure of finding the direction in latent W space of StyleGAN2-Ada for background removal and illumination modification.}
    \label{fig:nbcg}
\end{figure}

In this work, we extend the InterFaceGAN to generate passport like images by removing the background and illumination variations (as in real world scenarios the background situation often affects the illumination of faces). StyleGAN2-Ada learns how to generate photo-realistic images as well as the relation between the illumination and the environment (background). 
To generate the passport-like normal dataset, we firstly synthesize 500K images by randomly sampling from the latent space of StyleGAN2-Ada. To find the hyperplane in the latent W space that separates passport-like images and images in the wild, several criteria are defined in this work to select the positive set (passport like images) and the negative set (images in the wild). As most of the StyleGAN2-Ada generated images are well centered, we extract from $I_i$ (image i with size $n\times m$  transformed to gray level) one small rectangle from the top left region $L_i = I_i[0:h, 0:w]$ and one small rectangle from the top right region $R_i = I_i[0:h, m-w:w]$ (as shown in Figure \ref{fig:nbcg}). These two regions contain the background information that can be used to select the positive and negative set with the the average of the pixel values in these two regions $bcg\_m_i$ and standard deviation of the pixel values in these two regions  $bcg\_s_i$. A higher value of $bcg\_m_i$ indicates whiter background while a lower $bcg\_m_i$ indicates more uniform background. Hence, the generated images with the highest $bcg\_m_i$ and lowest $bcg\_s_i$ are selected as positive set (with label 1) and  the generated images with the lowest $bcg\_m_i$ and highest $bcg\_s_i$ are selected as negative set (with label -1). Then a linear SVM is trained to separate the positive set from the negative set in the latent W space. With the linear SVM, we can get the normal vector of the separation hyperplane $n_{bcg}$. Given any image $I_i$ generated from  its corresponding latent vector $W_i$, we can remove its background by editing the latent code: $W_{i\_edit} = W_i + \alpha \times n_{bcg}$.   

\begin{figure}
    \centering
    \includegraphics[width=0.49\textwidth]{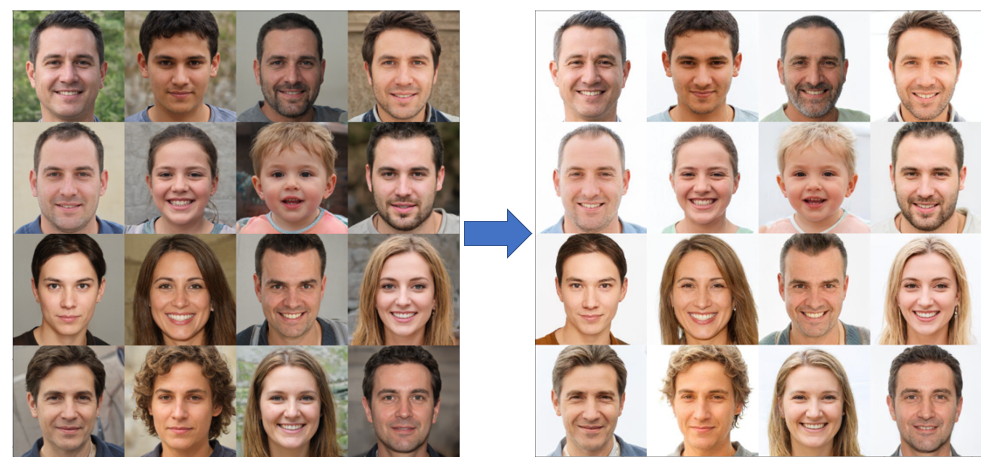}
    \caption{Examples of background removal and illumination modification by controlling the latent space of StyleGAN2-Ada. Images on the left side are generated by StyleGAN2-Ada through randomly sampled seeds. Images on the right are generated by using the proposed solution to remove the background. }
    \label{fig:bcgr}
\end{figure}

\begin{figure}
    \centering
    \includegraphics[width=0.49\textwidth]{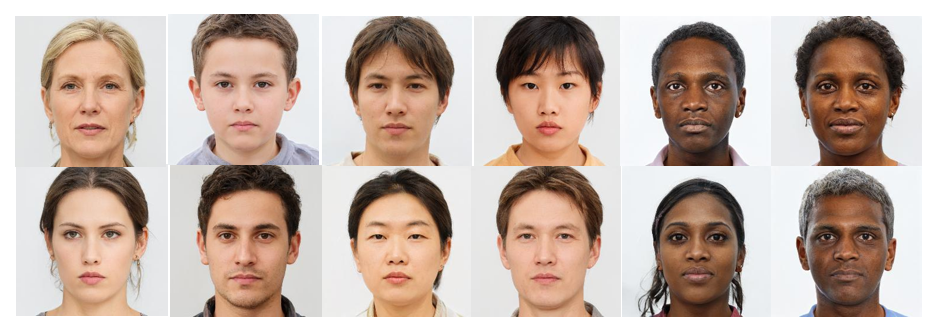}
    \caption{Examples of final generated normal passport-like dataset by controlling the latent space of StyleGAN2-Ada. }
    \label{fig:gendata}
\end{figure}

From the experimental results in Figure \ref{fig:bcgr}, we can tell that the proposed solution can remove the background and make the illumination more passport-like. However, there are still facial expression and small pose variations in the generated images.  
To deal with these limitations, we use InterfaceGAN to remove expressions like smiling. As most images generated by StyleGAN2-Ada (Trained on FFHQ dataset) have neither large pose variations nor harsh illuminations, we use the symmetric loss defined in Equation \ref{lsym} to select the most symmetric images. In this way, non-frontal pose and non frontal illumination images can be filtered. Finally, we also keep the skin color and gender balanced in the generated dataset to deal with the dataset bias in real world datasets. Some examples of the generated normal dataset are shown in Figure \ref{fig:gendata}.   

\section{Experiment setting}
\subsection{Training data}
For the non-normal dataset, we use the same CMU Multi-PIE dataset (the largest database for evaluating face synthesis and recognition in the controlled
setting) Setting-1 (12 poses and 20 illuminations of 150 identities for training) as in \cite{qian2019unsupervised} constrained setting for a fair comparison.
For the normal dataset, we use the generated passport-like image dataset described in the previous section in this work with balanced gender and skin color distribution. The total normal image dataset size is around 20K.

\subsection{Testing data}
As the non-normal training dataset is a constrained dataset from CMU Multi-PIE, we first evaluate the proposed StyleFNM model on the test data of CMU Multi-PIE Setting-1 (in the constrained environment), which includes 13 poses and 20 illuminations of 100 identities. To further study the generalization ability of the trained model to unconstrained environments, we also evaluate the proposed StyleFNM model on IJBA \cite{klare2015pushing}, which is one of the most challenging unconstrained face recognition benchmark datasets with uncontrolled pose variations, illumination variations, background variations, facial expression variations, blur and different accessories. Many of these variations in IJBA dataset are not present in our training data. IJBA contains both video frames and images from 500 subjects with 2,042 videos and 5,397 images that are split into 20,412 frames, 11.4 images and 4.2 videos per subject, captured from in-the-wild environment to avoid the near frontal bias, along with 10 random split protocols for evaluation of both verification (1:1 comparison) and identification (1:N search) tasks respectively.

\subsection{Experimental protocol}
 The pre-processing step (crop and alignment) of the training dataset is the same as in \cite{qian2019unsupervised}. The encoder part of the generator is a publicly-available pretrained ResNet-50 from VGG-Face2, whose parameters are fixed during the training and testing. The proposed StyleFNM is implemented in pytorch. The discriminators and the generator are trained iteratively to minimize the discriminator loss function and the
generator loss function in sequence with the Adam optimizer. The hyperparameters of the loss function are set to $\lambda_1 = 10$, $\lambda_2 = 0.1$, $\lambda_3 = 1$ based on a grid search with different parameter combinations.  The hyperparameters of the Adam optimizer is set as: $\alpha = 0.001$, $\beta_1 = 0$, $\beta_2=0.99$, $\epsilon = 10^{-8}$, following the suggestions in \cite{qian2019unsupervised}.  The proposed StyleFNM output size is 128x128 to make the inference speed faster for real world applications.

\section{Experimental results}

\subsection{Qualitative results}
We first show the visual results of the proposed StyleFNM. As mentioned in the previous sections, one of the limitation of the face normalization methods is being not realistic enough for human inspections, including distorted faces, incomplete ears, unrealistic asymmetric faces, etc. In this work, we try to deal with such limitations with the specific discriminator design.   
\subsubsection{Visual comparison on Multi-PIE dataset}

\begin{figure}
    \centering
    \includegraphics[width=0.45\textwidth]{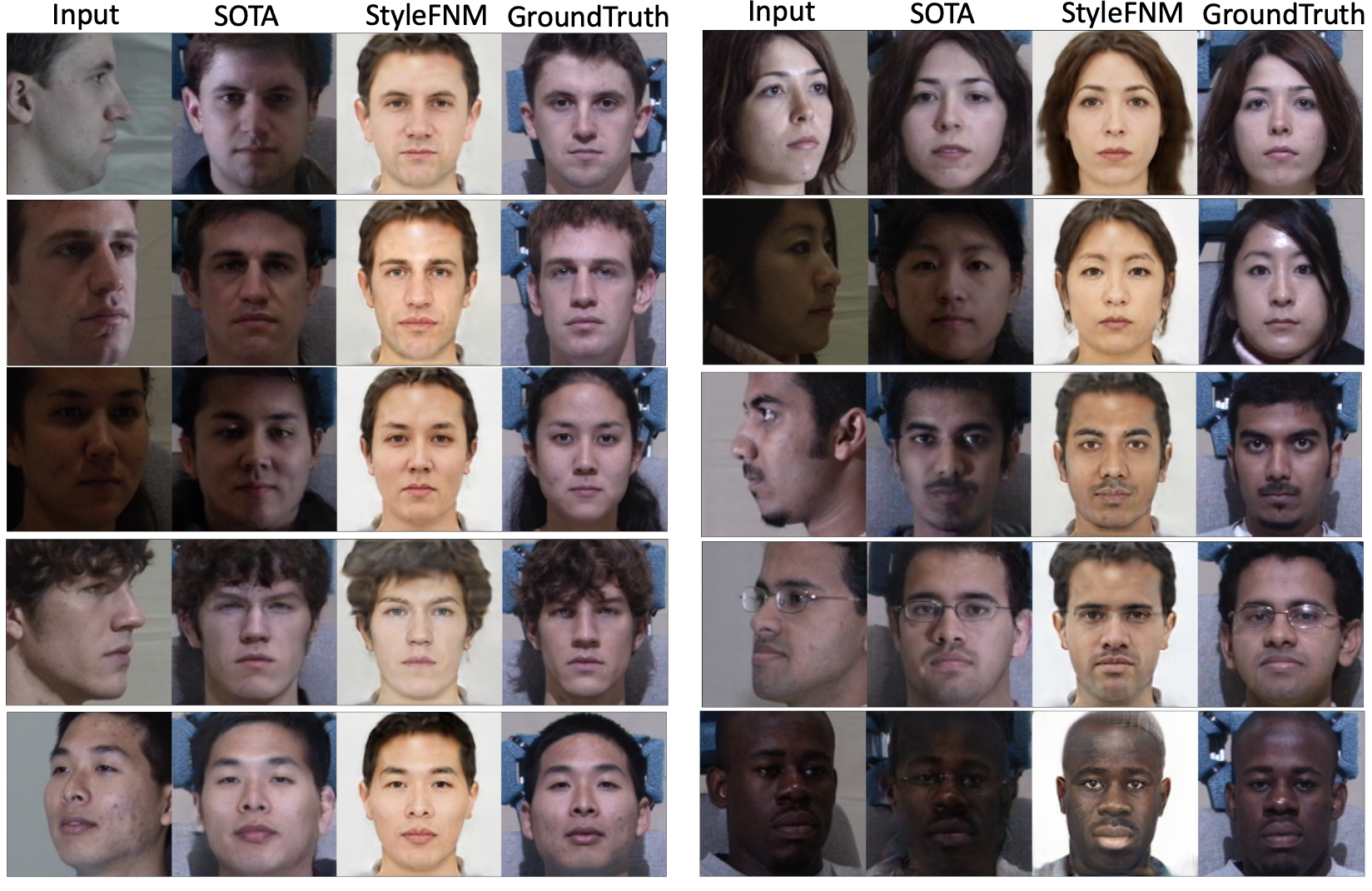}
    \caption{Examples of StyleFNM generated results on Multi-PIE dataset. Each pair contains 4 images:  input real image with different variations, FNM normalized output, StyleFNM normalized output and the ground truth normal image from left to right. }
    \label{fig:res1}
\end{figure}

  The StyleFNM face normalization results on CMU Multi-PIE dataset are shown in Figure \ref{fig:res1}. From this figure, we can see that the generated outputs all have passport-like white background, standard frontal illumination, standard neutral pose and neutral facial expression. This is a significant improvement comparied to FNM which can assign random illuminations and accessories like glasses to the outputs as shown in Figure \ref{fig:2} and Figure \ref{fig:res1}. From the visual comparison of StyleFNM outputs and the ground truth images, we can see that the identity information is better preserved than FNM, especially when the input image contains large pose or harsh illumination.  
  Overall, StyleFNM outputs have uniform frontal pose while FNM outputs have pose variations on pitch, roll and yaw.
  
  Generally speaking, when the illumination variation is removed, the facial identity features are visually more clear and easier to recognize. 
  From the examples at the bottom right in Figure \ref{fig:res1}, we show that when harsh illumination couples with dark skin color, it is difficult to see the facial identity even for human observers. The outputs generated by FNM not only adds glasses to the input but also make the input more difficult for face recognition task (for dark skin colors) while the StyleFNM outputs are much more fair and clear with well preserved identity information. 
 
 \begin{figure}
    \centering
    \includegraphics[width=0.49\textwidth]{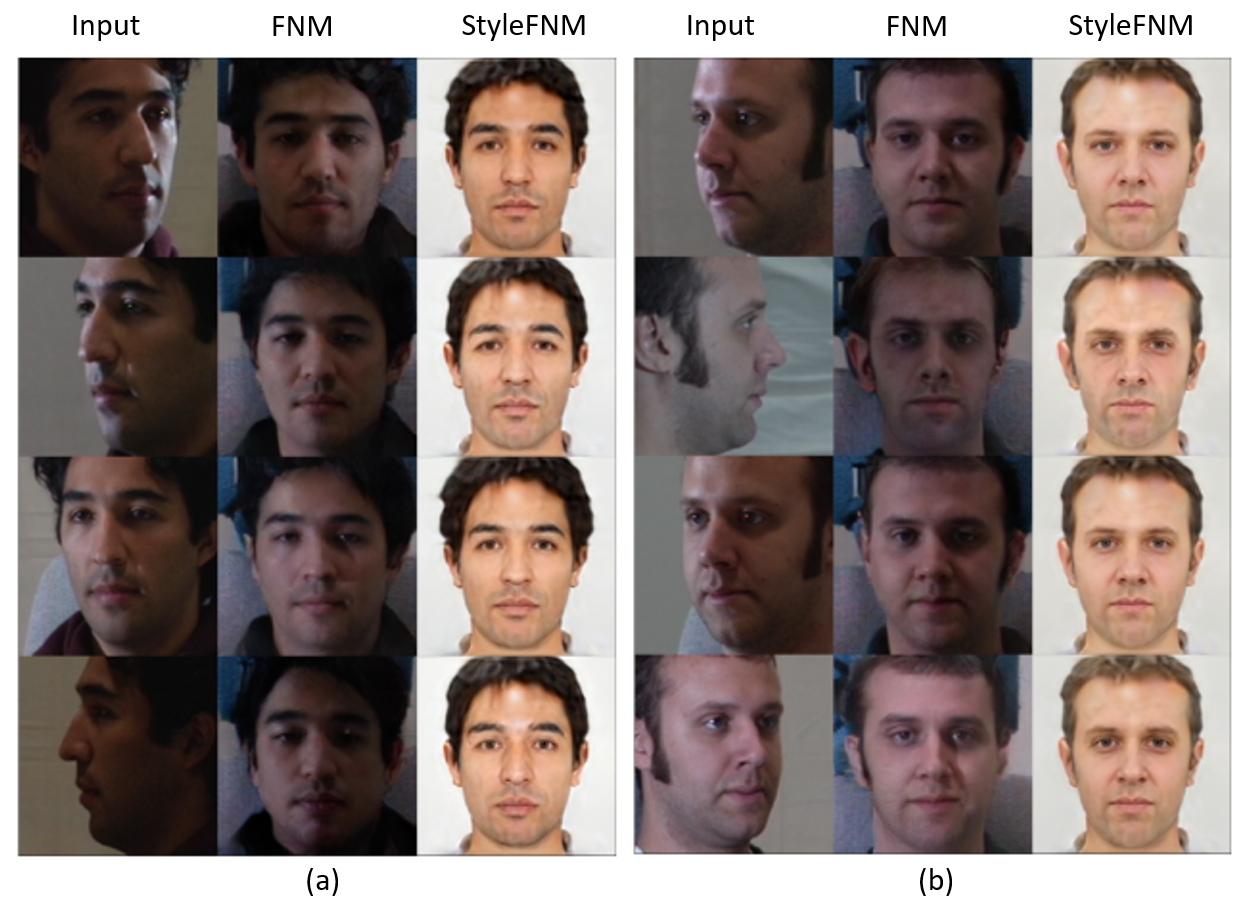}
    \caption{Two examples of the FNM and StyleFNM outputs from the same identity (each column) but with different pose and illumination variations.}
    \label{fig:comp1}
\end{figure}

 In order to show the model robustness as well as the effects of symmetric loss and our discriminator design, we show in Figure \ref{fig:comp1} the FNM and StyleFNM outputs from the same identity but with different pose and illumination variations. From the visualization results, it can be observed that: for the same identity, when the pose or the illumination changes, the FNM output changes too in terms of illumination, pose and facial identity. However, the outputs of the proposed StyleFNM are more stable and robust regardless of the changes in pose and illumination of the input. From Figure \ref{fig:comp1}, we can also see the effect of the symmetric loss proposed in this work: the StyleFNM outputs are more symmetric and stable than FNM outputs. In terms of discriminators, StyleFNM contains two more discriminators than FNM: left ear discriminator and right ear discriminator. It is shown in Figure \ref{fig:comp1} that StyleFNM generates more realistic ears than FNM.  
 
\subsubsection{Visualization of StyleFNM dealing with domain gap}

\begin{figure}
    \centering
    \includegraphics[width=0.49\textwidth]{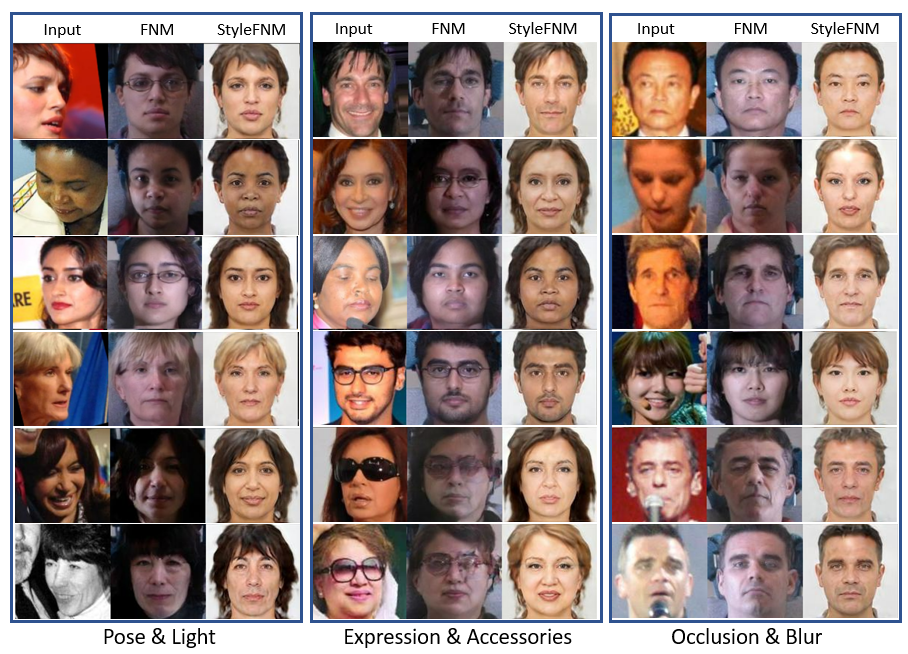}
    \caption{Visualization of StyleFNM dealing with domain gap: the input images are from IJBA dataset (unconstrained) while the StyleFNM is trained on constrained dataset. Examples of pose, light, expression, accessories, occlusion and blur are shown. For each pair of three images, the left image is the input image from IJBA, the middle one is the FNM output and the right one is the StyleFNM output.  }
    \label{fig:comp2}
\end{figure}

We also visualize the generalization ability of the trained StyleFNM model in Figure \ref{fig:comp2}. The proposed StyleFNM model is trained on constrained data (CMU Multi-PIE), which are collected in the constrained indoor environment. IJBA is an unconstrained dataset of face images in the wild, which contains much more illumination and background variations, and pose variations (row, pitch and yaw rotations) than CMU Multi-PIE dataset. Furthermore, IJBA contains variations that are never seen in the training data, including occlusion, blur and different accessories. Hence IJBA is a good dataset to test the generalization ability of our model on unseen input data with domain gaps.  From the results shown in Figure \ref{fig:comp2}, we can see that the proposed StyleFNM also has good performance on IJBA dealing with unseen variations such as sunglasses, motion blur and occlusion.

\subsection{Quantitative results}
In the previous section, we have shown qualitatively that the proposed StyleFNM can normalize most of the variations present in the training data (such as illumination, pose, facial expression, sunglasses etc.) as well as unseen variations including blur, occlusion and so on. In this section, the quantitative evaluation of the proposed solution is proposed by using StyleFNM as the pre-processing step for pretrained face recognition systems.
Following the experimental protocol in \cite{qian2019unsupervised}, pretrained Light CNN \cite{wu2018light} is selected as the face recognition model. To evaluate the performance of StyleFNM, the performance boost of Light CNN when using StyleFNM as a pre-processing step is measured. 

\subsubsection{Quantitative performance on constrained data}

\begin{table}
\centering
\caption{Rank-1 recognition rates (\%) across poses and illuminations
under Multi-PIE Setting-1.}\label{pie}
\begin{adjustbox}{width=0.49\textwidth} 
\begin{tabular}{ccccccc}
\multicolumn{1}{c}{Method} & \multicolumn{1}{l}{±90\textsuperscript{o}} & \multicolumn{1}{l}{±75\textsuperscript{o}} & \multicolumn{1}{l}{±60\textsuperscript{o}} & \multicolumn{1}{l}{±45\textsuperscript{o}} & \multicolumn{1}{l}{±30\textsuperscript{o}} & \multicolumn{1}{l}{±15\textsuperscript{o}} \\ 
\hline\hline
HPN \cite{ding2017pose} & 29.8 & 47.6 & 61.2 & 72.8 & 78.3 & 84.2 \\
c-CNN \cite{xiong2015conditional} & 47.3 & 60.7  & 74.4 & 89.0 & 94.1 & 97.0 \\
TP-GAN \cite{huang2017beyond} & 64.0 & 84.1 & 92.9 & 98.6 & 99.99 & 99.8 \\
PIM \cite{zhao2018towards} & 75.0 & 91.2 & 97.7 & 98.3 & 99.4 & 99.8 \\
CAPG-GAN \cite{hu2018pose} & 77.1 & 87.4 & 93.7 & 98.3 & 99.4 & 99.99 \\ 
\hline
Light CNN & 2.6 & 10.5 & 32.7 & 71.2 & 95.1 & 99.8\\
FNM+Light CNN & 55.8 & 81.3 & 93.7 & 98.2 & 99.5 & 99.9 \\ 

\makecell{StyleFNM  +Light CNN }& \textbf{86.2} & \textbf{95.5} & \textbf{99.0} & \textbf{99.8} & \textbf{99.9} & \textbf{100.0} \\
\hline
\end{tabular}

\end{adjustbox}
\end{table}

We first evaluate the performance of StyleFNM on the test data of CMU Multi-PIE setting 1. The rank-1 recognition rates on different poses are reported in Table \ref{pie}. Different state-of-the-art solutions are chosen for the comparison including HPN \cite{ding2017pose}, c-CNN \cite{xiong2015conditional}, TP-GAN \cite{huang2017beyond}, PIM \cite{zhao2018towards} and CAPG-GAN \cite{hu2018pose}. 
These methods either fine-tune the baseline (Light-CNN) on the Multi-PIE database, or are trained
with paired data and identity information. However, FNM and StyleFNM are trained with unpaired datasets and directly incorporated to face recognition model.

Generally speaking, the proposed StyleFNM achieves the best performance across all different poses, while the improvement is greater when the head pose is larger. For example, most methods have similar performances on the 15 degree head pose, while StyleFNM has an improvement of 9.1\% compared to CAPG-GAN on the 90 degree head pose. Compared to our baseline method FNM, StyleFNM has achieved significant improvement over all head poses, especially on large poses (30.4\% improvement on 90 degree head pose and 14.2\% improvement on 75 degree poses).

\subsubsection{Generalization performance on unconstrained data}
\begin{table}
\centering
\caption{The generalization performance comparison on IJBA benchmark. Results reported are the ’average±standard deviation’ over the 10 folds specified in the IJBA protocol. Symbol ’-’ indicates that the metric is not available for that protocol.}
\label{table:IJBA}
\begin{adjustbox}{width=0.49\textwidth} 

\begin{tabular}{ccccc} \hline
\multicolumn{1}{c}{\multirow{2}{*}{\begin{tabular}[c]{@{}l@{}}Method\\\end{tabular}}} & \multicolumn{2}{c}{Verification}                               & \multicolumn{2}{c}{Identification}                           \\ \cline{2-5}
\multicolumn{1}{l}{}                                                                  & \multicolumn{1}{l}{@FAR=0.01} & \multicolumn{1}{l}{@FAR=0.001} & \multicolumn{1}{l}{Rank-1} & \multicolumn{1}{l}{Rank-5}  \\ \hhline{-====}
OpenBR \cite{klare2015pushing}    & 23.6±0.9 & 10.4±1.4  & 24.6±1.1   & 37.5±0.8 \\
GOTS \cite{klare2015pushing}  & 40.6±1.4 & 19.8±0.8   & 43.3±2.1     &59.5±2.0  \\
DCNN  \cite{chen2016unconstrained}  & 78.7±4.3  & -          & 85.2±1.8    & 93.7±1.0  \\
DR-GAN \cite{tran2017disentangled}  & 77.4±2.7  & 53.9±4.3   & 85.5±1.5    & 94.7±1.1 \\
FF-GAN  \cite{yin2017towards}  & 85.2±1.0  & 66.3±3.3   & 90.2±0.6    & 95.4±0.5   \\ \hline
LightCNN  & 82.7±2.0  & 67.4±2.2   & 84.5±1.7    & 92.6±0.9   \\ 
FNM+LightCNN   & 93.0±1.4  & 84.0±2.8   & 93.4±1.2    & 96.7±0.7     \\

\textbf{StyleFNM+LightCNN}       & \textbf{ 94.6±1.1}    & \textbf{ 87.7±2.6}  &\textbf{ 94.9±1.2}   & \textbf{97.4±1.1}   \\ \hline
\end{tabular}
\end{adjustbox}
\end{table}

In order to study the generalization performance of the proposed StyleFNM model trained on constrained dataset, we also perform the quantitative evaluation on the unconstrained dataset IJBA.  Similar to the evaluation setting on CMU Multi-PIE data, we also use LightCNN as the baseline model and compare StyleFNM to FNM (also trained on constrained dataset) as well as other state-of-the-art methods.  From the results in Table \ref{table:IJBA}, we can see that the proposed StyleFNM trained on constrained dataset can generalize well to unconstrained dataset IJBA. Particularly, StyleFNM with LightCNN achieves  11.9\% improvement at FAR 0.01 and 20.3\% improvement at FAR 0.001 on face verification, 10.4\% improvement at Rank-1 and 4.8\% improvement at Rank-5 on identification. Compared to FNM, StyleFNM also achieves great improvements in terms of generalization ability.  For example, compared to FNM, StyleFNM  achieves  1.6\% improvement at FAR 0.01 and 3.7\% improvement at FAR 0.001 on face verification.

\subsection{Fairness study}
\begin{figure}
    \centering
    \includegraphics[width=0.49\textwidth]{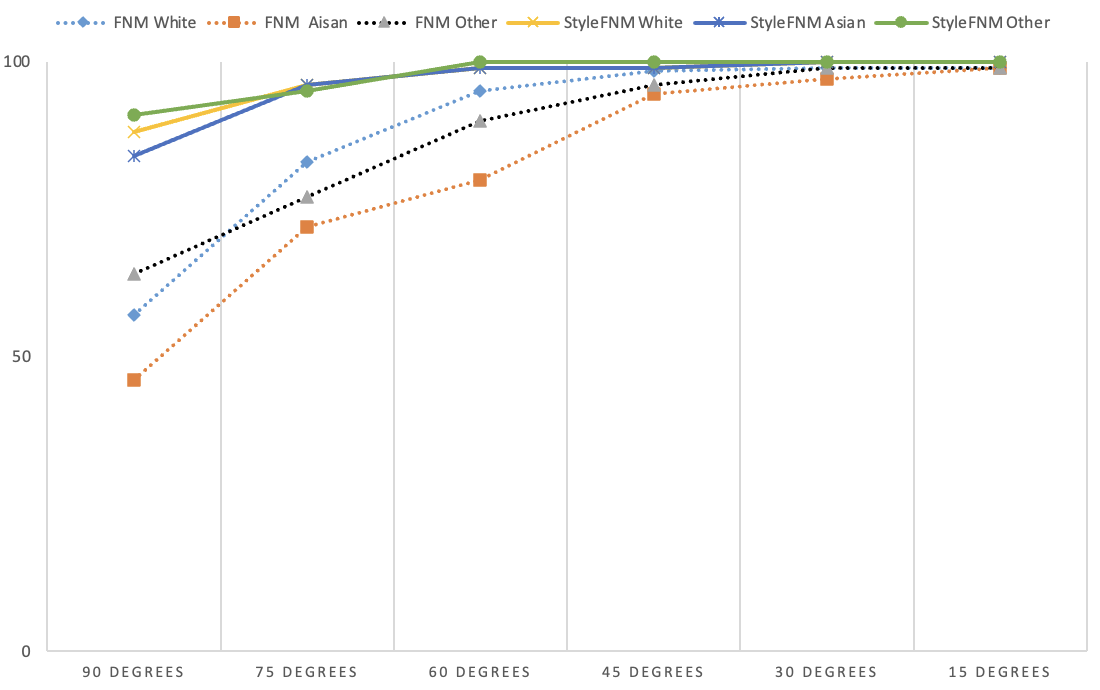}
    \caption{Rank-1 recognition rates (\%) of StyleFNM VS FNM across different ethnicities and across different poses under Multi-PIE Setting-1.} 
    \label{fig:bias}
\end{figure}

In section 3, we have introduced the passport-like normal image data generation method. To deal with the skin color bias (the proposed solution can also be used to deal with other biases such as gender bias, age bias etc.) which exists in many state-of-the-art solutions as well as in many real world datasets, the generated normal image dataset contains balanced skin colors so that trained StyleFNM would be less biased. We evaluate also the performance of StyleFNM across different skin colors on the test data of CMU Multi-PIE setting 1. The results are shown in Figure \ref{fig:bias}. Generally speaking, StyleFNM has similar face recognition performances across different skin colors, especially for poses within 75 degrees, while FNM has bigger performance gap among different skin colors. 

\subsection{Ablation study}
\subsubsection{Component analysis} 
\begin{figure}
    \centering
    \includegraphics[width=0.4\textwidth]{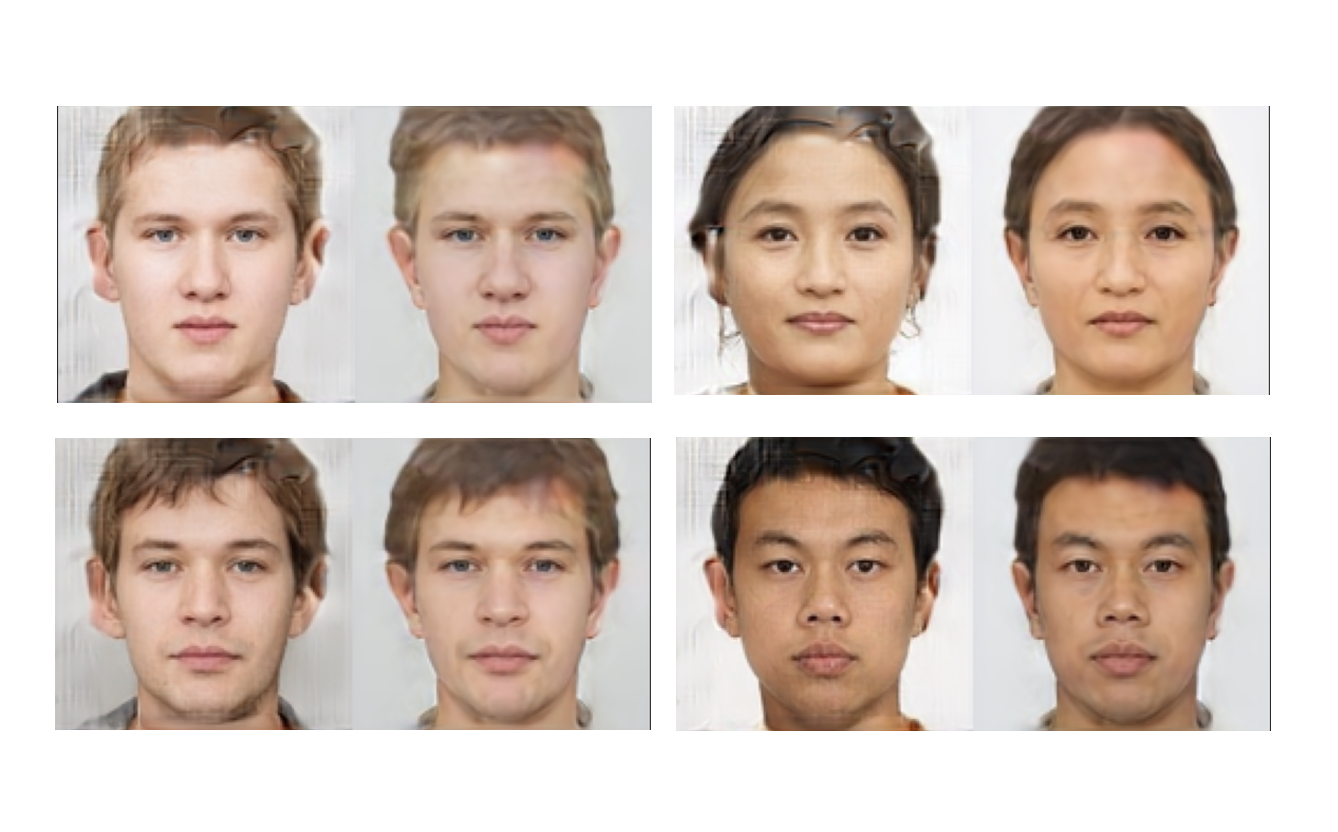}
    \caption{Visual comparison between StyleFNM (old discriminators) outputs (on the left side of each pair of images) and StyleFNM (new discriminators) outputs (on the right side of each pair of images).} 
    \label{fig:ablation}
\end{figure}

In this work, we have shown that the proposed StyleFNM can generate more realistic normalized face images and make the face recognition task easier both for machines and human inspectors with three new components compared to FNM: the generated normal passport-like training data, the style based generator and the discriminator design. Extra experiments are done to study the effects of each component. 
To study the contribution of the generated passport-like normal image dataset, we  re-train the original FNM with the new generated passport-like normal image dataset. From the result shown in Table \ref{ablation} second row, it can be seen that FNM trained with our generated new normal dataset outperforms the original FNM, especially for 90 degrees head poses (around 10\% improvement). This finding indicates that removing more variations to transform the input image to the standard passport format can make the face recognition task easier. On top of the new generated dataset, we then add the new generator to FNM. The result is shown as StyleFNM (old discriminators) in 3rd row of Table \ref{ablation}. It can be seen that the new generator design contributes to most of the improvements.  This finding is in line with our hypothesis in previous sections that the generator capacity of the original FNM is limited. By comparing StyleFNM (old discriminators) with StyleFNM with our new discriminator design, we can see that the  improvement in terms of face recognition performance is not as big as the other two components (new normal data and new generator design). However, the new discriminator design is still very important. In section 2, we pointed out that one limitations of existing face normalization methods is the unrealistic outputs, especially in ears region. The new discriminator design is to make sure that the generated output is more realistic. The visual comparison between StyleFNM with old discriminators and StyleFNM with new discriminators are shown in Figure \ref{fig:ablation}. It can be seen that the StyleFNM with new discriminators focuses on different details of the face, which makes its output look more realistic. 

\begin{table}
\centering
\caption{Rank-1 recognition rates (\%) across poses and illuminations
under Multi-PIE Setting-1 for ablation studies. }\label{ablation}
\begin{adjustbox}{width=0.49\textwidth} 
\begin{tabular}{ccccccc}
\multicolumn{1}{c}{Method} & \multicolumn{1}{l}{±90\textsuperscript{o}} & \multicolumn{1}{l}{±75\textsuperscript{o}} & \multicolumn{1}{l}{±60\textsuperscript{o}} & \multicolumn{1}{l}{±45\textsuperscript{o}} & \multicolumn{1}{l}{±30\textsuperscript{o}} & \multicolumn{1}{l}{±15\textsuperscript{o}} \\ 
\hline\hline
FNM+Light CNN & 55.8 & 81.3 & 93.7 & 98.2 & 99.5 & 99.9 \\ 
\hline
\makecell{FNM (new normal data) \\ +Light CNN }  & 65.4 & 83.5 & 93.8 & 98.0 & 99.4 & 99.8  \\ \hline
\makecell{StyleFNM (old discriminators)\\ +Light CNN }& {84.5} & {94.4} & {98.7} & \textbf{99.8} & \textbf{99.9} & \textbf{100.0} \\
\hline
\makecell{StyleFNM  \\ +Light CNN }& \textbf{86.2} & \textbf{95.5} & \textbf{99.0} & \textbf{99.8} & \textbf{99.9} & \textbf{100.0} \\
\hline
\end{tabular}

\end{adjustbox}
\end{table}

\subsubsection{Face normalization speed at inference time} 
In real world applications, the face normalization model is used as pre-processing step for face recognition models. Hence, the speed of face normalization model needs to be fast for its utility. We evaluate the inference speed of StyleFNM on one Nvidia Tesla T4 GPU. Given an input image, it takes around 0.026 second to pass through the encoder and generator design to get the normalized image. With 38 frames per second inference speed, the proposed StyleFNM is acceptable for many real world applications.

\section{Conclusion and future works}
In this paper, we summarize the limitations of the state-of-the-art face normalization methods and propose solutions to deal with them. Firstly, we propose to deal with the dataset bias by controlling a pretrained GAN  to generate passport-like standard images as the normal dataset so that we can remove most kinds of variations. With the proposed method, we can also control the demographic distribution to ensure the final dataset is more inclusive and fair.  
Then, we propose a novel face normalization model StyleFNM to translate the input image to the standard passport-like format using the style based generator along with the proposed loss functions and discriminators. The proposed StyleFNM can generate good quality photo-realistic outputs while keeping the identity unchanged. Experimental results show that the proposed StyleFNM can improve the performance as well as fairness of face recognition systems significantly. StyleFNM also generalizes well to other datasets despite the presence of domain gaps. However, there are still several limitations in this work. 
Only one expert network is used in this work, while multiple expert networks can be combined together as in \cite{cao2018improve, cao2019random}  to improve the robustness and the generalization ability of the proposed solution.
As StyleFNM focuses more on facial identity, the hairstyle is not well preserved. Other details such as eye color and face structure preservation could also be improved in the future.    

The proposed face normalization system can have other applications in the real world scenarios. In a real world face recognition system, the gallery photos are passport-like while the probe photos are often from live cameras with different environments and backgrounds. However, most existing datasets have the gallery and probe photos from the same domain/environment due to the data acquisition protocol. With the proposed face normalization system, we can generate the passport-like gallery photos (paired with the  probe in real world environments) and simulate a more realistic scenario to test the face recognition system performance. The proposed system can also be used to aid the human inspectors dealing with surveillance cameras to facilitate the face recognition task. 

\bibliographystyle{ieeetr}
\bibliography{sample}
\end{document}